\documentclass[letterpaper, 10 pt, conference]{ieeeconf}  

\IEEEoverridecommandlockouts                              

\usepackage{graphics} 
\usepackage{epsfig} 
\usepackage{mathptmx} 
\usepackage{times} 
\usepackage{amsmath} 
\usepackage{amssymb}  
\usepackage{graphicx} 
\usepackage{xcolor}
\usepackage{algorithm} 
\usepackage{algpseudocode}
\usepackage{multicol}
\usepackage{hyperref}
\usepackage{url}
\usepackage{multirow}
\usepackage{adjustbox}
\usepackage{etoolbox}
\usepackage[font=small]{caption}
\usepackage{paralist}
\usepackage{flushend}
\usepackage[symbol]{footmisc}

\DeclareMathOperator*{\argmin}{arg\,min}

\graphicspath{{imgs/}}
\title{\LARGE \bf
Imagine2Servo: Intelligent Visual Servoing with Diffusion-Driven Goal Generation for Robotic Tasks
}

\author{Pranjali Pathre$^{1}$, Gunjan Gupta$^{1}$, M. Nomaan Qureshi$^{2}$, Mandyam Brunda$^{1}$, \\
Samarth Brahmbhatt$^{3}$,
K. Madhava Krishna$^{1}$
\thanks{$^{1}$Robotics Research Center, IIIT Hyderabad, India}%
\thanks{$^{2}$Carnegie Mellon University, USA.}
\thanks{$^{3}$Intel Labs}}

\begin{document}
\maketitle
\thispagestyle{empty}
\pagestyle{empty}

\begin{abstract}

Visual servoing, the method of controlling robot motion through feedback from visual sensors, has seen significant advancements with the integration of optical flow-based methods. However, its application remains limited by inherent challenges, such as the necessity for a target image at test time, the requirement of substantial overlap between initial and target images, and the reliance on feedback from a single camera. This paper introduces Imagine2Servo\footnote[2]{Project Page: \url{https://brunda02.github.io/RRC/}}, an innovative approach leveraging diffusion-based image editing techniques to enhance visual servoing algorithms by generating intermediate goal images. This methodology allows for the extension of visual servoing applications beyond traditional constraints, enabling tasks like long-range navigation and manipulation without pre-defined goal images. We propose a pipeline that synthesizes subgoal images grounded in the task at hand, facilitating servoing in scenarios with minimal initial and target image overlap and integrating multi-camera feedback for comprehensive task execution. Our contributions demonstrate a novel application of image generation to robotic control, significantly broadening the capabilities of visual servoing systems. Real-world experiments validate the effectiveness and versatility of the Imagine2Servo framework in accomplishing a variety of tasks, marking a notable advancement in the field of visual servoing.
\end{abstract}
\section{INTRODUCTION}

The visual servoing problem~\cite{538972} involves the challenge of controlling the motion of a robot by utilizing feedback from visual sensors, typically cameras, to adjust its actions in real-time. This process entails the robot's ability to interpret visual data to determine its relative position and orientation with respect to target objects or locations within its environment. The core objective is to enable the robot to perform precise movements or reach specific goals by continuously comparing the current visual scene against a desired configuration or outcome. This approach requires sophisticated algorithms for image processing and control theory to bridge the gap between visual perception and mechanical action, thereby allowing the robot to adapt its movements based on the visual feedback it receives.

Recently, there has been a lot of progress in optical flow-based visual servoing methods~\cite{rtvs, katara2021deepmpcvs, argus2020flowcontrol, harish2020dfvs, 9550239}. These methods are shown to be highly precise in reaching their goals with some guarantees of convergence. However, the utility of visual servoing has remained limited due to major limitations common to all servoing algorithms: 1) They necessarily require a goal image during test time. This makes it quite tough for visual servoing algorithms to be applied in real-world navigation or manipulation since if we already have a map of the environment, there are better ways to reach the goal pose than through a target image. 2) Visual Servoing cannot work if there is not much overlap between the initial and target image. 3) Visual Servoing can only accommodate feedback from a single camera. 

Solving each of these problems can greatly enhance the utility of visual servoing methods. For example, solving the problem of final image generation based on the skill the robot is executing can make servoing quite useful for real-world tasks. Imagine a drone mounted with a monocular camera trying to cross a door. The robot will first have to visualise the approximate position of the door just before it crosses it; then, it will have to visually servo to the imagined image and then apply a simple hardcoded skill to cross the door. This pattern of imagining a goal, servoing to the goal and applying a hardcoded skill can be repeated for many skills, both in navigation and manipulation. Another good example of a skill we can solve using servoing is the `reaching' skill, which is a part of several manipulation tasks. For instance, take the example of `unplugging the charger', where the robot has to `reach' a particular grasping pose before applying a hardcoded policy.

\begin{figure}[!t]
    \centering
    \includegraphics[width=\columnwidth]{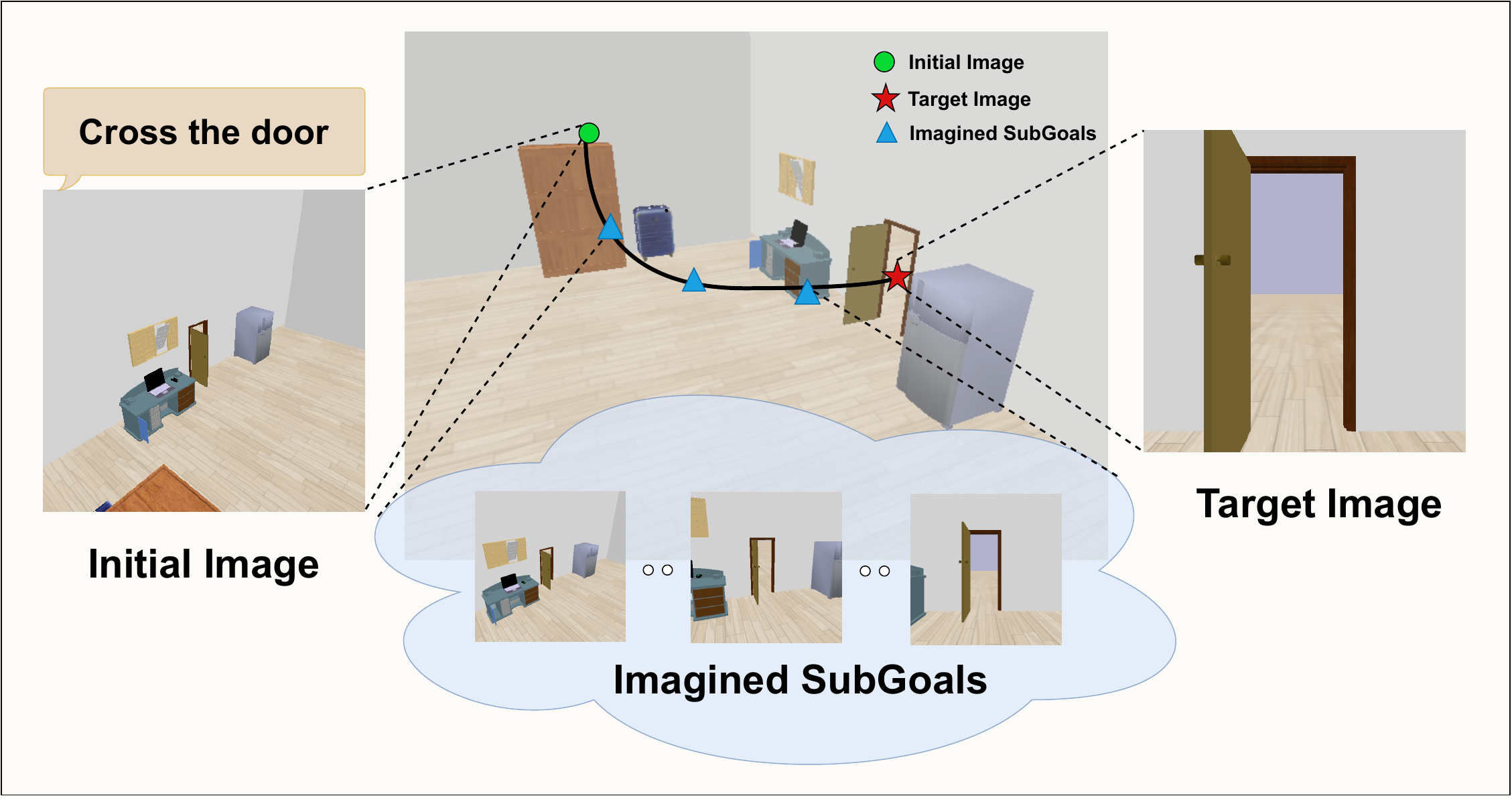}
    \caption{Given a single eye-in-hand camera input and the language instruction to perform a task, we introduce \textbf{Imagine2Servo}, a pipeline that generates intermediate goal images, which are then used by the Image-Based Visual Servoing (IBVS) controller to reach a target location. We show the application of our pipeline to long-range navigation as well as manipulation tasks.}
    \label{fig:teaser}
\end{figure}

In this paper, we leverage the recent advancements in diffusion-based image editing to provide a much-needed and major update to servoing algorithms. Our contribution can be stated as follows:
\begin{itemize}
\item  We first introduce the notion of final image generation for servoing, grounded in the skill we are trying to execute. We then introduce our Imagine2Servo pipeline, which takes as input the current view of the camera and outputs a subgoal for the servoing algorithm to reach.
\item We show that the Imagine2Servo pipeline can be used to solve the problems where traditional servoing fails even with a target image in place. This is especially pertinent when the initial image and target image don't have much overlap. 
\item We demonstrate how we can incorporate feedback from multiple cameras in servoing. We show a thorough evaluation on a variety of tasks in different simulators such as Habitat~\cite{habitat19iccv}, RLBench~\cite{rlbench}, and PyBullet~\cite{pybullet}.
\item
We show real-robot results of our method to further demonstrate its efficacy.
\end{itemize}

\section{RELATED WORKS}
\textbf{Diffusion Models}: Recent years have witnessed a surge in the development of diffusion models, with advancements occurring at an astonishing pace. Building upon the seminal works of \cite{ho2020denoising, sohldickstein2015deep, dhariwal2021diffusion, rombach2022highresolution}, diffusion models have been successfully applied to a wide range of tasks, including text-driven image generation \cite{ramesh2022hierarchical, saharia2022photorealistic}, depth estimation\cite{saxena2023monocular, ke2023repurposing}, image inpainting~\cite{lugmayr2022repaint} and video editing~\cite{molad2023dreamix}. Recently, there has also been a lot of work in applying diffusion models for robot learning~\cite{chi2023diffusion, ajay2023conditional, janner2022planning, ze20243d}. In this paper, we are particularly interested in works which use smart data collection strategies to train diffusion models for image editing~\cite{brooks2023instructpix2pix, kawar2023imagic, meng2022sdedit}. We investigate how these image editing models, specifically instruct-pix2pix~\cite{brooks2023instructpix2pix}, trained using diffusion techniques, can be leveraged to produce high-quality images that effectively represent the desired goals for the servoing models to achieve.

\textbf{Visual Servoing}: Deep learning-based visual servoing approaches have emerged as an alternative to traditional methods reliant on hand-crafted features. In the last few years, optical flow-based visual servoing algorithms~\cite{rtvs, katara2021deepmpcvs, argus2020flowcontrol, harish2020dfvs, 9550239} have gained a lot of traction. They use off-the-shelf optical flow networks~\cite{teed2020raft, fischer2015flownet} as a feature. These algorithms are quite robust and have been applied to a variety of tasks, from large building avoidance~\cite{sankhla2022flow} to dynamic grasping~\cite{10354813}. However, they still require a target image to servo, which limits their applicability in a lot of real-world tasks where the goal image is not available. In contrast, our framework generates the subgoal for the servoing to reach. We show in our results that not only our skill-grounded Imagine2Servo framework can beat traditional servoing algorithms, even in cases where the servoing methods have a final image to go to, but can also be applied to other tasks which fall in the reach and execute category.


A work notably related to ours is SuSIE~\cite{black2023zeroshot}, which employs the InstructPix2Pix~\cite{brooks2023instructpix2pix} architecture in tandem with reinforcement learning (RL) policies for solving manipulation tasks. While SuSIE is a notable advancement, our research focuses significantly on updating servoing models. Unlike RL methods, which are computationally intensive and sample-inefficient, our servoing controller is broadly applicable without fine-tuning for specific datasets. This makes it more deployable across diverse environments, providing a more efficient and flexible solution. Another pertinent work is Robotap~\cite{vecerik2023robotap}, which employs keypoint tracking combined with a visual servoing algorithm for certain manipulation challenges. Robotap is limited by its reliance on well-defined key points, making it ineffective for tasks like navigating through a door where key points are hard to identify. In contrast, our model's versatility overcomes these limitations, handling a wider range of tasks. Visual foresight models~\cite{finn2017deep, ebert2018visual} are another line of work which are similar to our primary motivation; however, learning a model in image space is challenging. We use a simpler, model-free approach that combines diffusion-based foresight with servoing, enhancing servoing's utility and creating a versatile pipeline for solving various tasks.
\section{METHOD}

\begin{figure*}[t]
    \centering
    \includegraphics[width=\textwidth]{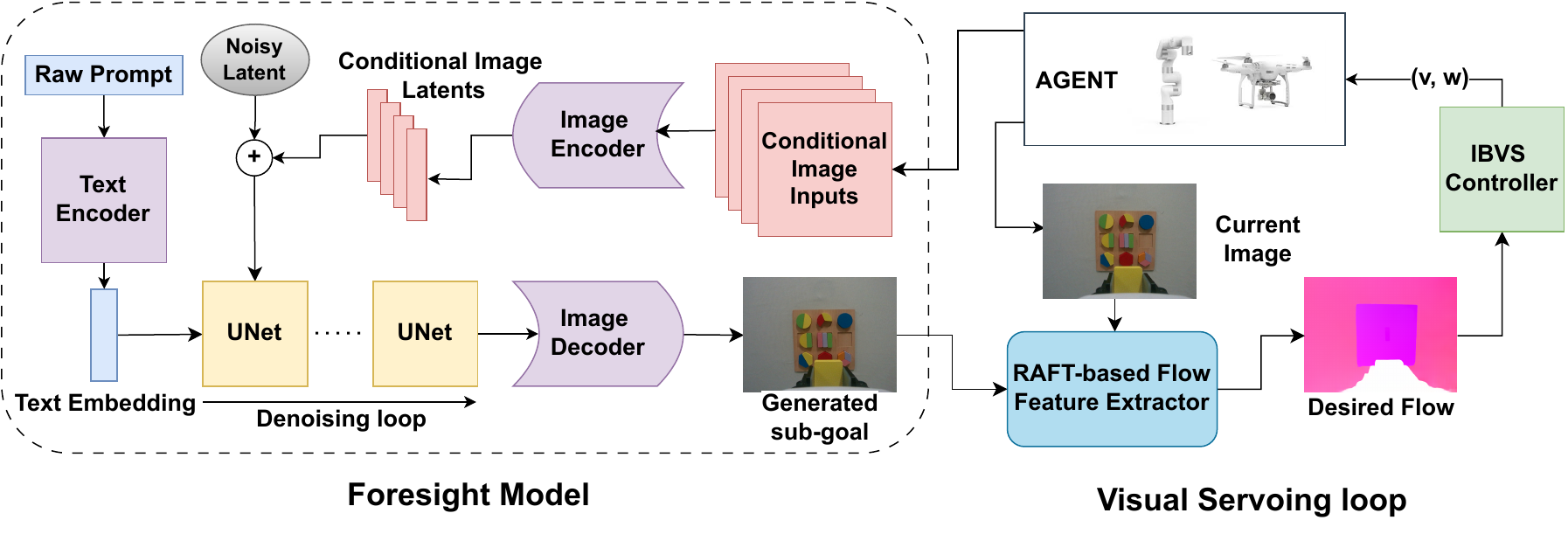}
    \caption{Our method alternates between generating sub-goals using a diffusion-based foresight model and executing actions with a flow-based IBVS controller. The foresight model, conditioned on the current monocular eye-in-hand camera input and any available scene observations, translates images to guide the task. The RTVS visual servoing controller predicts 6 DOF actions to reach each intermediate goal, repeating the loop until the task is completed.}
    \label{fig:pipeline}
\end{figure*}

\subsection{Problem Formulation}


At any given moment, our system receives an RGB image $I_t$ from the robot's camera sensor, along with a textual prompt $P$ that specifies the task to be accomplished. Our objective is to accurately predict and implement the necessary control commands, denoted as [$v_t, \omega_t$], to fulfil the task described in the prompt.  Tasks vary, ranging from navigating through a doorway to disconnecting a charger. Note that in traditional servoing algorithms, the final image $I_g$ is usually provided to perform the task. This greatly limits the utility of the visual servoing models. To address this challenge, we divide our approach into two distinct phases. Initially, our framework is designed to conceptualize a subgoal $I_g$ based on the task and current camera feedback $I_t$. Next, we aim to attain this subgoal image $I_g$ by employing a servoing algorithm. In the forthcoming sections, we will elaborate on the components of our Imagine2Servo framework. We begin by introducing our foresight model for sub-goal generation in Section \ref{foresight} followed by the mechanics of our servoing framework \ref{servoing}, which facilitates reaching the subgoal. We then describe our overall framework in \ref{overall} and training details in \ref{training}. Our overall pipeline is summarised using Fig. \ref{fig:pipeline}.



\begin{algorithm*}
\caption{Imagine and Servo, Test-Time execution}\label{alg:cap}
\begin{algorithmic}[1]
\Require Test time $t$, language prompt \textit{P}, current observation $I_t$, Image foresight model $p(I_g^{*} | I_t, $\textit{l}$)$, IBVS controller  $\phi(I_t, I_g^{*})$, convergence constant for foresight model $\epsilon_{p}$, convergence constant for IBVS controller $\epsilon_{\phi}$
\State $t \gets 0$, $dist \gets \infty$
\State Sample $I_g \sim p(I_g | I_t, $P$)$ 
\Comment{\textit{Generate next intermediate goal given the language instruction and the current observation}}
\While {$dist \geq \epsilon_{p} $}  \Comment{\textit{Check if the last generated goal is the same generated goal.}}
\State $I_g^{last} \gets I_g$
\While { $||I_g - I_t|| \leq \epsilon_{\phi}$} \Comment{\textit{Convergence criterion for IBVS}}
\State $ \mathcal{F}_t := $ predict-target-flow ($I_g, I_t)$ \Comment{\textit{Predict the flow using the flow network}}
\State $ \mathcal{L}_t := $ calculate-interaction-data ($\mathcal{F}(I_g, I_t)$) \Comment{\textit{Calculate the interaction matrix}}
\State Execute $v_{t+1} := $ relative-pose-network ($\mathcal{L}_t, \mathcal{F}_t$) \Comment{\textit{Predict and execute action on the robot}}
\State $I_{t+1} \gets $ robot-observation \Comment{\textit{Get the current observation after action execution}}
\State $t \gets t+1$ 
\EndWhile
\State Sample $I_g \sim p(I_g | I_t, $l$)$ 
\Comment{\textit{Generate the next intermediate goal.}}
\State $dist \gets ||I_g^{last} - I_g|$
\EndWhile
\end{algorithmic}
\end{algorithm*}

\subsection{Imagine2Servo Architecture}


\subsubsection{Visual foresight model}
\label{foresight}

We see the task of generating the next subgoal for the servoing controller as editing the pixels of the current input image. Given the current image $I_t$ and task description $P$, we aim to generate the subgoal image $I_g$ using our foresight model $p(I_g|I_t, P)$. To achieve this task, we use Instructpix2pix~\cite{brooks2023instructpix2pix}, an image editing framework, though any other image editing model can be used. Instructpix2pix gives a good initialisation of the denoising model conditioned on the text prompt and the input image. We utilize the pretrained weights of this model and fine-tune it further to fit our task of learning visual foresight. 

One significant benefit of this approach is the ability to integrate auxiliary camera data into the denoising process. This enhancement proves invaluable in scenarios where a solitary camera's perspective is inadequate for generating the necessary subgoal. An illustrative example of this is when the camera is affixed to the end-effector of a manipulator; in such instances, auxiliary image information $I_{\text{aux}}$ from alternative viewpoints can significantly augment the planning of subgoals. Our empirical results demonstrate that the inclusion of auxiliary views markedly enhances the algorithm's efficacy, particularly in manipulation tasks. This methodology facilitates the incorporation of multi-camera feedback into the servoing algorithm, a capability that traditional servoing frameworks do not possess.

\subsubsection{Image Based Visual Servoing Controller}
\label{servoing}
In our approach, we employ RTVS~\cite{rtvs} (Real-Time Visual Servoing) as the Image-Based Visual Servoing (IBVS) controller to reach the subgoals predicted by the foresight module. We use the RTVS servoing algorithm without any fine-tuning to the newer environments. RTVS is an iterative algorithm. At every timestep $k$, it seeks to optimize camera motion velocities (linear and angular) $\mathbf{v}_k$ to minimize the difference between the velocity-induced optical flow $\widehat{\mathcal{F}}(\mathbf{v}_k)$, and $\mathcal{F}(I_k, I_g)$, the optical flow of the current image $I_k$ to the given subgoal image $I_g$.

\begin{equation}
    \mathbf{v}^*_k = \argmin ||\widehat{\mathcal{F}}(\mathbf{v}_k) -  \mathcal{F}(I_k, I_g)||_2^2
\end{equation}

In RTVS, this optimization is performed by inference on a pre-trained motion generation network (see~\cite{rtvs} for details). $\widehat{\mathcal{F}}(\mathbf{v}_k)$ is calculated using the interaction matrix $L(Z_t)$ (described in Equation \ref{eq:interaction}), which relates the camera velocity to the corresponding flow in the image plane.

\begin{equation}
\label{eq:interaction}
L(Z_t) = \begin{bmatrix}
    -1/Z_t & 0 & x/Z_t & xy  & -(1+{x}^2) & y \\
    0 & -1/Z_t & y/Z_t & 1+{y}^2  & -xy & -x
    \end{bmatrix}
\end{equation}

This matrix is a function of the depth $Z_t$ and the image coordinates $(x, y)$, encapsulating the geometric relationship between motion in the robot's operational space and its projection onto the image plane. We used the FlowDepth trick described in \cite{rtvs} to use flow as a proxy for depth, thereby preventing the need for depth information in our pipeline. The cycle of optimizing the camera velocity, moving the camera by that velocity, and then computing the optical flow from the new observed image to the subgoal image is repeated until convergence, i.e. until the photometric error between the current observed image and the subgoal image is less than a threshold.

\subsection{Overall Imagine2Servo Framework}
\label{overall}
Our Imagine2Servo framework is described in Algorithm~\ref{alg:cap}. Given a task prompt $P$, the algorithm leverages the current observation $I_t$ to initiate a process that dynamically guides the robot towards achieving the specified task. This is facilitated by using an Image Foresight Model $p(I_g | I_t, P)$~\ref{foresight}, which generates a series of intermediate goal images $I_g$. These images act as visual targets for the robot to pursue, bridging the gap between the high-level task description and the robot's moment-to-moment actions.

At each time step, the algorithm assesses the alignment between the generated goal image and the current observation using a predefined convergence criterion $\epsilon_p$. If the current goal is deemed to have been reached, the model generates a new goal image based on the updated observation and the initial task prompt $P$. This process repeats, with the robot making progress towards the final objective through a series of intermediate steps.

The action execution phase is managed by an IBVS controller $\phi(I_t, I_g^*)$, which adjusts the robot's actions to minimize the difference between its current state and the target state represented by the goal image as described in Section~\ref{servoing}. This methodology allows the robot to navigate and manipulate objects within its environment effectively, translating high-level language prompts into a sequence of actionable visual goals. Our algorithm enhances adaptability and efficiency by continually updating goals with real-time feedback, offering a novel integration of language-based instructions with robotic control.

%
%
\subsection{Training and implementation details}
\label{training}
Imagine2Servo is trained on input-output image tuples with a corresponding text description. We use the pre-trained weights of InstructPix2Pix~\cite{brooks2023instructpix2pix} and fine-tune their model. We train our model on a single 12 GB RTX 3090 GPU for 30,000 training steps with a learning rate of $5 \times 10^{-5}$ and 500 learning rate warm-up steps. We use a batch size of 4 with 4 gradient accumulation steps. We use images of dimensions of 256 x 256, and the rest of the hyperparameters are similar to InstructPix2Pix. At test time, we use 100 inference steps with 1.5 and 7.5 as our image and text guidance parameters, respectively.
\section{EXPERIMENTS AND ANALYSIS}
\label{sec:experiments}

We evaluate the performance of Imagine2Servo for two different tasks: 1) \textit{Door crossing}. This tests the ability of our algorithm to perform long-range servoing. Experiments are conducted in two simulation environments: PyBullet \cite{pybullet} and Habitat \cite{habitat19iccv}. 2) \textit{Reaching}. This tests the applicability of our algorithm to manipulation tasks. Experiments are performed both in simulation (RLBench~\cite{rlbench}) and on a real robot setup.

\subsection{Dataset} \label{dataset}
We collect our dataset in a variety of environments like Pybullet~\cite{pybullet}, Habitat~\cite{habitat19iccv} and RLbench~\cite{rlbench}. 1) \textit{Door crossing}: Using the door samples from the PartNet-Mobility dataset~\cite{partnet}, we generate a dataset of 720 video sequences of a drone crossing the door, with 9 frames per sequence. The captured video sequences resemble that of a drone trying to cross a door frame. From every consecutive image pair of a trajectory, we sample input-output image tuples. The text prompt expected by the InstructPix2Pix backbone, which describes the task being performed, is set to ``cross the door". We diversify scene elements during video creation to generate synthetic scenes that closely resemble real-world environments, randomizing initial poses, lighting, wall textures, and floors in each video. 2) \textit{Reaching}: We choose 17 tasks from RLbench~\cite{rlbench} and collect 100 demonstrations for every variation of a task. Every task in RLBench is a composition of reaching and manipulating actions. So, we collect data to learn reaching by discarding manipulation actions like grasping from the demonstrations. The text prompt is set to the textual description of the task provided by RLBench. 

\subsection{Baselines} \label{sec:experiments:baselines}
We compare our Imagine2Servo framework with the following approaches:
\begin{itemize}
    \item \textbf{RTVS} \cite{rtvs} is an MPC-based formulation with multi-step ahead prediction. We provide the privileged real final image as the target image and report the success rate.
    \item In \textbf{Pose-Diffusion}, we implement a simplified version of the Diffusion Policy \cite{diffusionpolicy} to predict the next action conditioned on the current wrist camera observation.
    \item In the \textbf{Cam-Axis} baseline, we implement a simple controller that moves the camera along the camera axis from the initial pose. 
\end{itemize}
We compare these baselines to two versions of our algorithm. \textbf{Imagine2Servo-Single-View} uses the current eye-in-hand camera view as the only conditional image input. In \textbf{Imagine2Servo-Multi-View}, we train Imagine2Servo to use additional input from the overhead camera as the conditional input to the foresight model, along with the eye-in-hand monocular camera input.

\subsection{Quantitative Comparison}
\label{sec:experiments:quantitative}


We measure the performance of all algorithms in terms of success rate. For door crossing, the camera is modelled as if it were mounted on a cylindrical drone of radius 15 cm. If the camera crosses the door, the trial is marked successful. For reaching tasks, we report the linear and angular errors as well as the success rate. The angular error is measured as the norm of quaternion difference. If the linear error $<$ 3 cm and angular error $<$ 0.03, the trial is considered successful. For both tasks, we also check for collision after every controller step and, upon collision, declare the trial a failure. We show success rate comparisons in table \ref{table:quantitative}.

Imagine2Servo-Single-View outperforms all the other baselines with a significant improvement in navigation tasks. It attains 90\% and 95\% success rates, respectively, in navigating doors in PyBullet \cite{pybullet} and Habitat-Sim \cite{habitat19iccv} environments, with 45\% absolute improvement over RTVS\cite{rtvs}, which employs a single target view to servo to the target. Imagine2Servo-Single-View outperforms Pose-Diffusion, which directly predicts actions instead of images. By reasoning in image space, our model leverages a powerful diffusion backbone trained on millions of images, an advantage Pose-Diffusion lacks. 
In RLBench reach tasks, Imagine2Servo-Multiview achieves a 75\% success rate, outperforming other baselines and surpassing Imagine2Servo-Single-View by 10\% with the help of additional camera views. It also achieves the lowest linear and angular error, enabling precise alignments crucial for manipulation tasks.

\begin{table}
\begin{adjustbox}{width=\columnwidth}
\begin{tabular}{cccccc}
\hline
\multirow{2}{*}{} & PyBullet & Habitat-Sim & \multicolumn{3}{c}{RLBench} \\ 
\cline{2-6} 
& \multicolumn{2}{c}{Success Rate}  & 
\begin{tabular}[c]{@{}c@{}}Success\\ 
Rate\end{tabular} & 
\begin{tabular}[c]{@{}c@{}}Trans.\\ Error
\end{tabular} &
\begin{tabular}[c]{@{}c@{}}Rot.\\ Error
\end{tabular} \\ 
\hline
RTVS\cite{rtvs} & 0.45 &  0.5 & 0.25 & 0.18 & 0.38\\
Pose-Diffusion & 0.4 & 0.2 & 0.2 & 0.27 & 0.45\\
Cam-Axis & 0.2 & 0.1 & 0.1 & 0.39 & 0.71\\ 
\hline
Imagine2Servo-Single-View (Ours) &  \textbf{0.9} & \textbf{0.95} & 0.65 & 0.09 & 0.08\\
Imagine2Servo-Multi-View (Ours)  & - & - & \textbf{0.75} & \textbf{0.03} & \textbf{0.02}\\ 
\hline
\end{tabular}
\end{adjustbox} 
\caption{\textbf{Quantitative results}: We benchmark the two different versions of our network along with three baselines (refer to Section~\ref{sec:experiments}). We report the average success rate for all tasks, and additionally, 6 DOF pose error for manipulation tasks in RLBench\cite{rlbench}. Imagine2Servo achieves the best success rate, demonstrating its applicability in different tasks.}
\label{table:quantitative}
\end{table}

\begin{figure}
    \centering
    \includegraphics[width=\columnwidth]{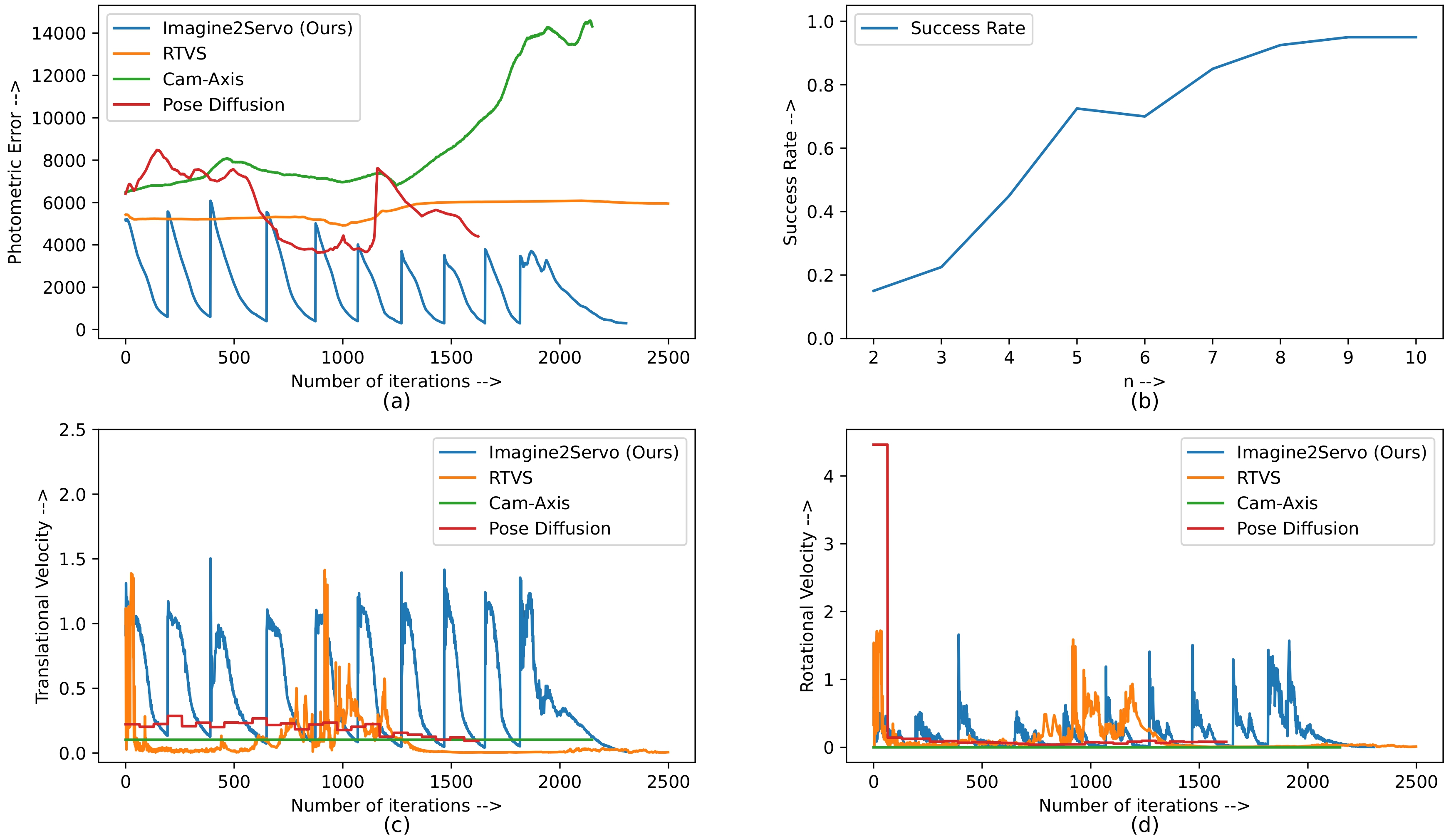}
    \caption{\textbf{(a)} We compare the photometric error convergence for each baseline. \textit{Imagine2Servo} converges to each of the subgoals (where the error becomes close to zero) while other methods fail to converge. \textbf{(b)} Ablation study to understand the effect of change in \textit{n}, the sampling frequency used to generate the training set on the success rate (refer to Section~\ref{sec:experiments:ablations}). \textbf{(c)} Evolution of the magnitude of the translational velocity in each step. \textbf{(d)} Evolution of the magnitude of the rotational velocity in each step.}
    \label{fig:quantitative_fig}
\end{figure}

To further analyze the performance of our model, we perform a comparative study of photometric error convergence (fig. \ref{fig:quantitative_fig}(a)) and evolution of translational and rotational velocities (fig. \ref{fig:quantitative_fig}(c),(d)) in each step. These experiments are performed on HM3D \cite{hm3d} scene (e.g. 2 in fig. \ref{fig:qualitative_nav}). The photometric error of Imagine2Servo smoothly converges without any oscillations to below $300$ (for all sub-goals), while other baselines fail to converge beyond the error of $4000$. We also observe a bounded velocity profile in Imagine2Servo while sudden accelerations are observed in RTVS~\cite{rtvs}. The initial trajectory in the Cam-Axis baseline heads toward the goal but ultimately collides with the front wall. Pose-Diffusion converges initially but fails to generate appropriate velocity commands near the target, leading to stalling and divergence. In contrast, Imagine2Servo decelerates smoothly as it nears the sub-goal image, resulting in a consistent motion profile. 


\subsection{Qualitative Results}
\label{sec:experiments:qualitative}

\begin{figure}[t]
    \centering
    \includegraphics[width=0.94\columnwidth]{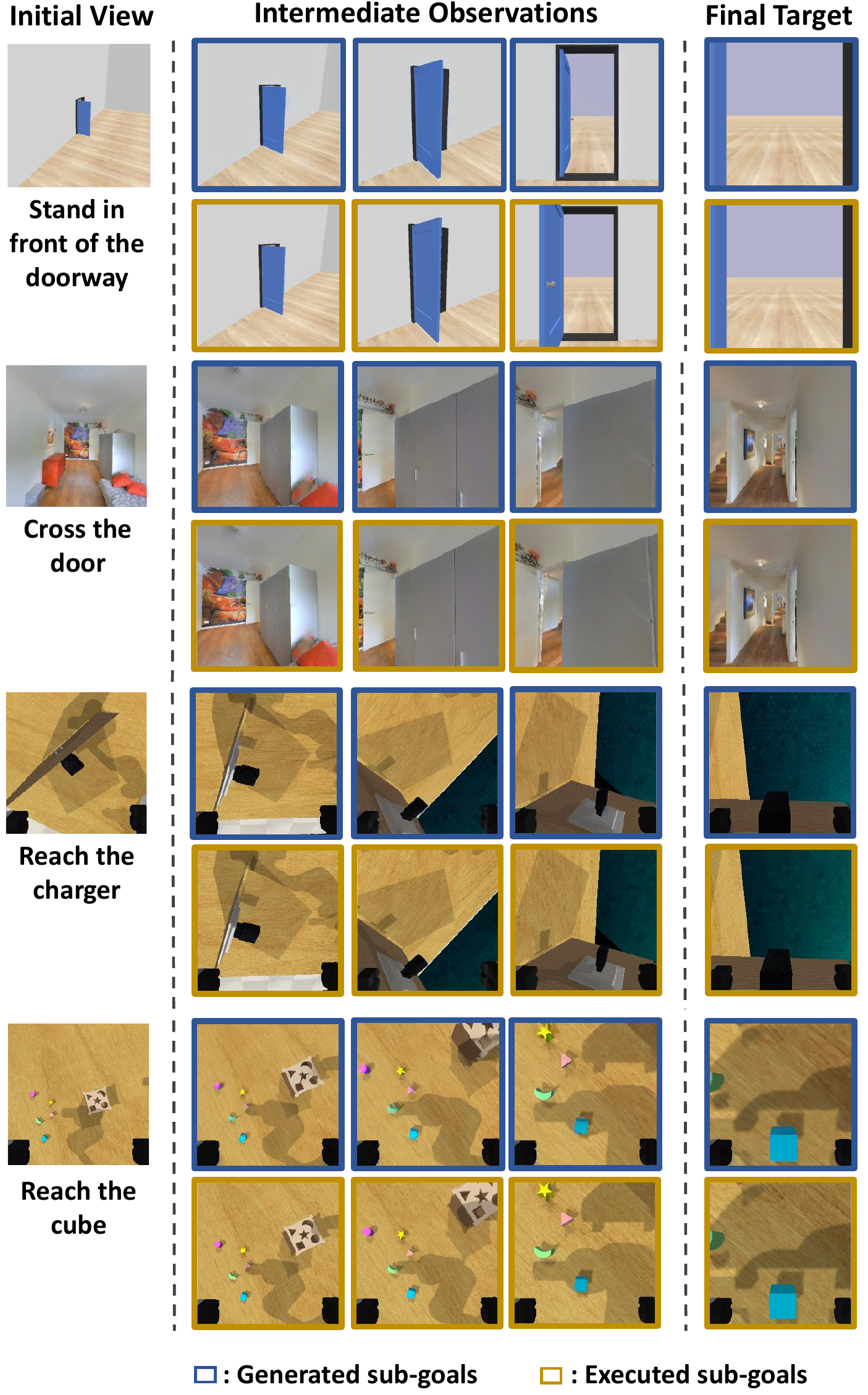}
    \caption{We visualize the foresight model generated sub-goals (blue border) and the pose reached (yellow border) after the execution of actions by the controller for the task of navigation through a door in PyBullet~\cite{pybullet} and Habitat-Sim~\cite{habitat19iccv}, and manipulative actions in RLBench~\cite{rlbench}. Our model successfully converges for large translations and rotations even when multiple objects are present in the scene.}
    \label{fig:qualitative_nav}
\end{figure}

Fig.~\ref{fig:qualitative_nav} shows qualitative results for door crossing and reaching, respectively.
To evaluate the reaching task, we pick 3 RLBench tasks with varying complexities. With sub-goal guidance, even if the object is obliquely visible in the image (e.g. 3 in fig. \ref{fig:qualitative_nav}), Imagine2Servo-Single-View achieves precise final alignment of the end-effector. It accurately interprets text input to perform the needed action, even when multiple objects are in the scene. For the shape sorter task, the foresight model can correctly reason between different objects (e.g. 4 in fig. \ref{fig:qualitative_nav}) to generate the next sub-goal. 

\begin{figure}
    \centering
    \includegraphics[width=\columnwidth]{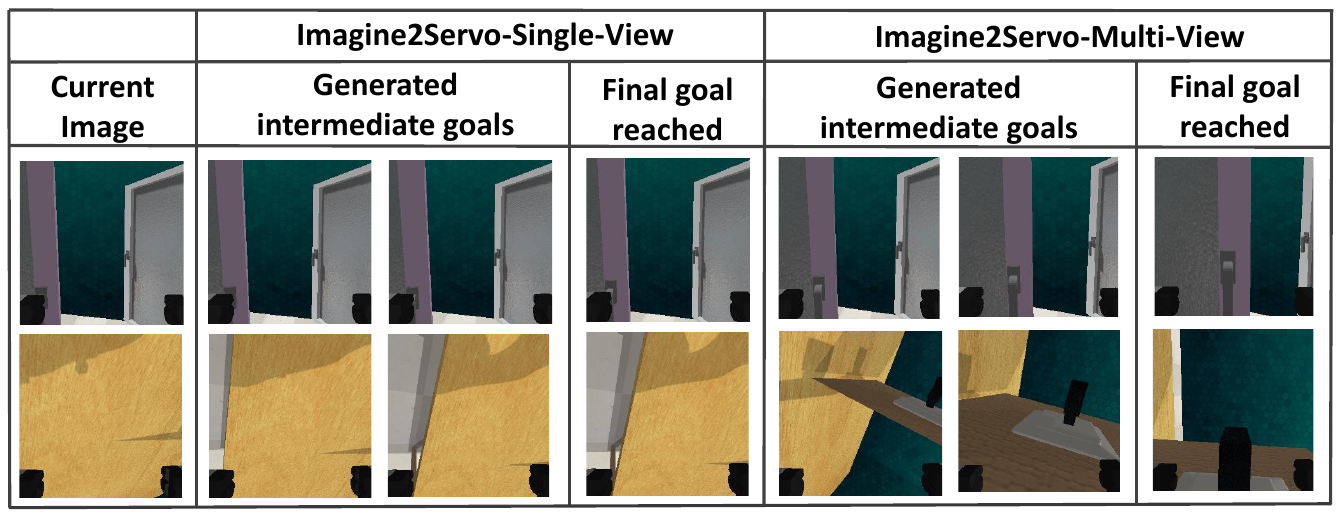}
    \caption{We qualitatively compare the performance of \textit{Imagine2Servo-Single-View} and \textit{Imagine2Servo-Multi-View} for the task of ``reach the window" (row 1) and ``reach the charger" (row 2) in RLBench \cite{rlbench}. 
    When the target is not visible or is occluded behind other objects in the initial image, \textit{Imagine2Servo-Single-View} fails to converge, whereas \textit{Imagine2Servo-Multi-View} generates sub-goals that converge to the final target.
    }
    \label{fig:multiview}
\end{figure}

Fig \ref{fig:multiview} summarizes the improvements of Imagine2Servo-Multi-View over the Imagine2Servo-Single-View baseline. Imagine2Servo-Multi-View correctly predicts the next sub-goal when the target object is not visible (row 2 in fig \ref{fig:multiview}) in the current view or is occluded behind other objects (row 1 in fig \ref{fig:multiview}) in the scene. In such cases, Imagine2Servo-Single-View diverges as the foresight model fails to generate the next relevant sub-goal. Using the auxiliary information from the overhead camera view, the multi-view model correctly generates the next intermediate view and guides the controller towards the target. 

\subsection{Real World}
\label{sec:experiments:realworld}

\begin{figure}
    \centering
    \includegraphics[width=0.95\columnwidth]{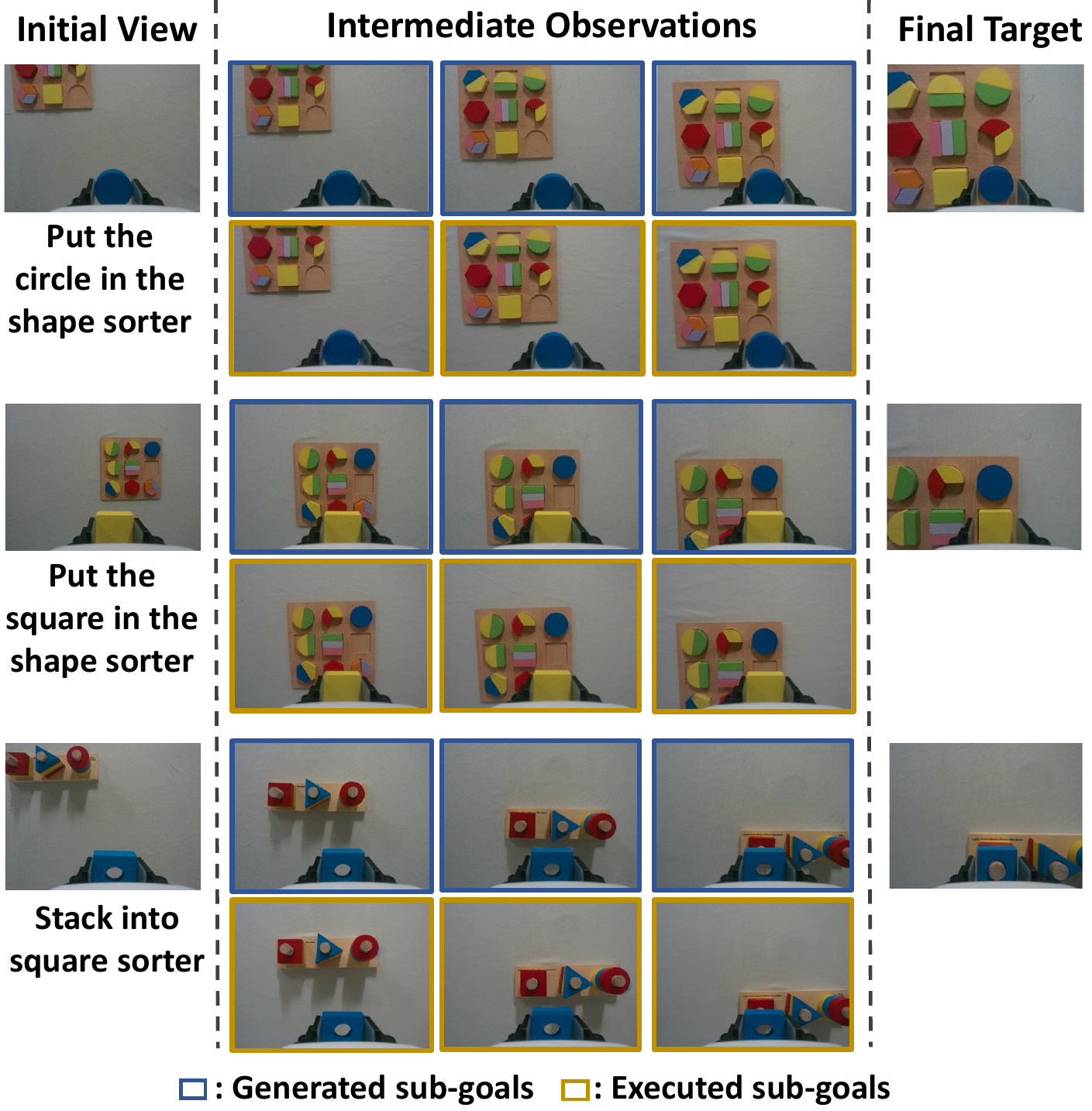}
    \caption{For real-world tasks, we visualize the generated sub-goals (blue border) and executions (yellow border) for the task of shape sorting and shape stacking.}
    \label{fig:realworld}
\end{figure}

We set up manipulation experiments in the real world on UFACTORY xARM7. We demonstrate the following tasks: \textbf{1.} Placing a shape in the shape sorter \textbf{2.} Shape stacking. We generate 300 trajectories with $15$ frames per trajectory for each shape and finetune our model by sampling input-output image tuples with a language prompt explaining the task. Unseen poses and randomized positions are used during test time. Our model performs reasonably well, achieving success rates of 8/10 for task 1 and 7/10 for task 2. Please see the supplementary material for videos. Imagine2Servo captures the semantic information in the scene and correctly sorts circles and squares in the respective shape sorters (rows 1, 2 in fig \ref{fig:realworld}). In task 2 it performs exceptionally by performing precise alignment which is important to successfully stack the object. Even when the stacker and the sorter are partially visible in the initial image, the foresight model generates relevant sub-goals and converges to the final target. Despite the actuation noise in the real-world setup, our method adapts well and converges successfully in the execution of the task.

\begin{figure}
    \centering
    \includegraphics[width=0.95\columnwidth]{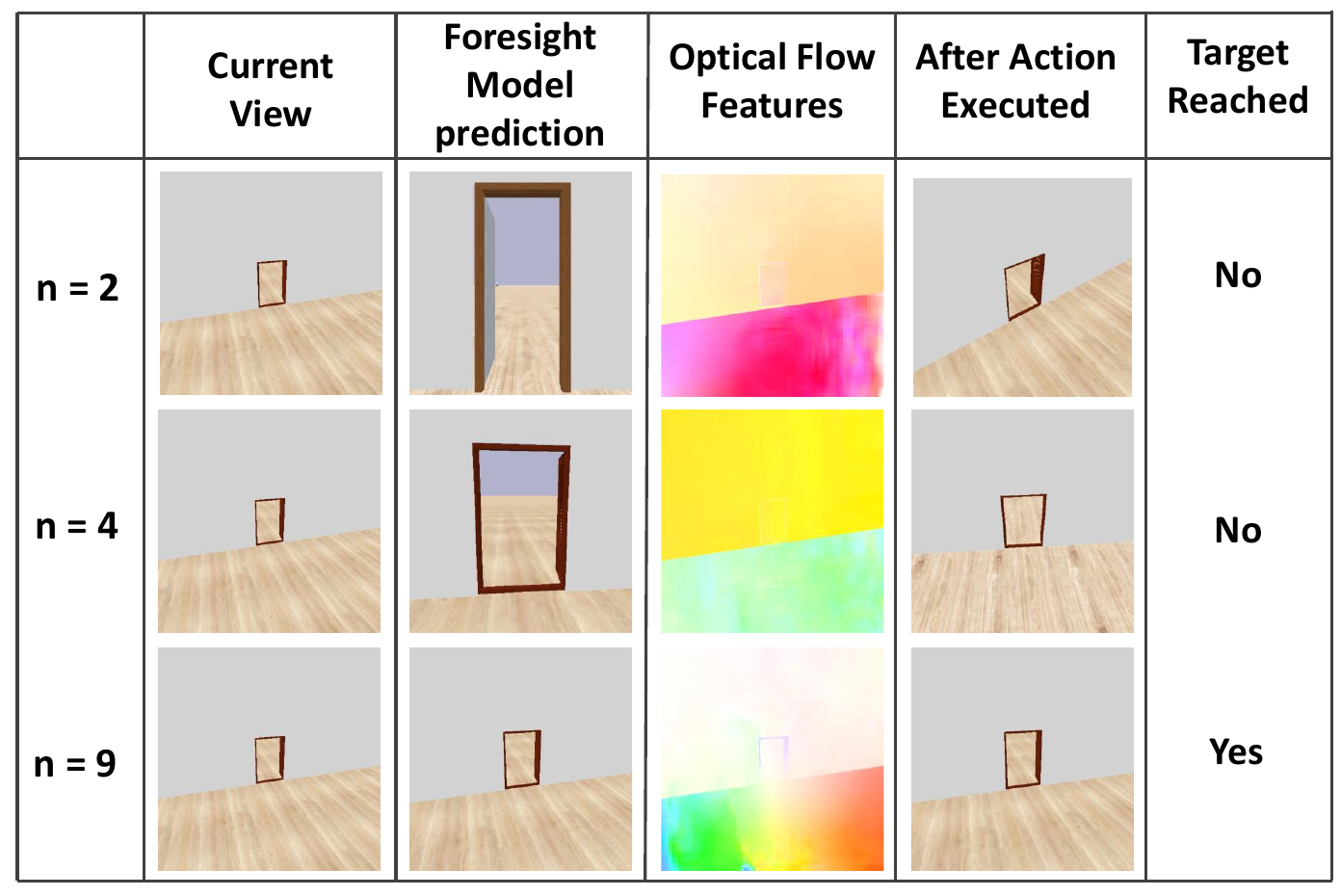}
    \caption{Results of qualitative comparison of sub-goal generations and IBVS controller convergence with the change in the value of \textit{n} (\textit{n} is the sampling frequency used to generate the training set. Refer to \ref{sec:experiments:ablations} for more details). We show results for the task of reaching the door in PyBullet \cite{pybullet}.}
    \label{fig:ablation}
\end{figure}

\subsection{Ablation Study}
\label{sec:experiments:ablations}


We perform ablation experiments to comprehend the advantage of using sub-goals in image-based servoing. At test time, we expect the flow-based IBVS controller to proficiently match the features of the current and the goal images generated by the foresight model. Optical flow calculations are constrained by the motion between two consecutive image frames. Therefore foresight model should be trained to predict the next sub-goal, which would give us good features for servoing. We study the convergence of Imagine2Servo as we uniformly sample \textit{n} frames per trajectory to form the training set $\textit{S} = \{I_1, ..., I_n\}$.


As \textit{n} increases, the optical flow between the current and the generated image become more prominent and lead to an increase in success rate (fig. \ref{fig:quantitative_fig}(b)). For $\textit{n} = 2$, we directly predict the final goal image without any sub-goals. The target image is often incorrect as the foresight model fails to generate the correct target view. Even if the target view is correct, the IBVS controller fails to converge to the desired target location. With the increase in the value \textit{n}, from n=4, the foresight model output becomes more grounded, yet it fails to converge due to the lack of good flow features. For \textit{n} = 9, the IBVS controller converges to the final target with balanced in-range sub-goal generations by the foresight model. Predicting sub-goals instead of a final goal grounds the foresight model and strengthens the controller by helping it compensate for errors in sub-goals (e.g. 2 in fig \ref{fig:qualitative_nav}).
\section{CONCLUSIONS}

We introduced Imagine2Servo, a novel framework that significantly extends the capabilities of visual servoing by generating intermediate goal images, enabling robust task execution in a range of environments without predefined goal imagery. This approach effectively addresses key limitations of traditional visual servoing methods, such as the requirement of goal images, the challenges posed by limited overlap between initial and target views, and the dependency on single-camera feedback. Results demonstrate the potential of Imagine2Servo in enhancing robotic navigation and manipulation capabilities through the innovative use of diffusion-based image editing for subgoal generation. Imagine2Servo sets a new benchmark for visual servoing, paving the way for more autonomous and adaptable robotic systems.
\section{ACKNOWLEDGMENT}

The authors acknowledge the support provided by MeitY, Govt. of India, under the project ``Capacity building for human resource development in Unmanned Aircraft System (Drone and related Technology)".


\bibliographystyle{IEEEtran}
\bibliography{references}


\end{document}